\documentclass[10pt,twocolumn,letterpaper]{article}

\usepackage{cvpr}
\usepackage{times}
\usepackage{epsfig}
\usepackage{graphicx}
\usepackage{amsmath}
\usepackage{amssymb}

\usepackage[T1]{fontenc}
\usepackage[utf8]{inputenc}
\usepackage[american]{babel}
\usepackage[font=small,labelfont=bf]{caption}
\usepackage{enumitem}

\usepackage{threeparttable}
\usepackage{multirow}
\usepackage{longtable}
\usepackage{rotating}
\usepackage{tabularx}

\usepackage{bbm}
\usepackage{bm}
\usepackage{pifont}
\usepackage{amssymb}
\usepackage{arydshln}
\usepackage{blindtext}

\usepackage{amsfonts}

\usepackage{booktabs}
\usepackage{color}
\usepackage[rgb,dvipsnames]{xcolor}
\definecolor{Olive_Green}{rgb}{0.0, 0.55, 0.0}
\usepackage[pagebackref=true,breaklinks=true,letterpaper=true,colorlinks,bookmarks=false, citecolor = Olive_Green]{hyperref}

\cvprfinalcopy 

\def\cvprPaperID{
	****
} 

\ifcvprfinal\fi   
\begin{document}

\title{Deeper and Wider Siamese Networks for Real-Time Visual Tracking
\vspace{-0.5em}
}


\author{Zhipeng Zhang\\
University of Chinese Academy of Sciences\&CASIA\\
{\tt\small zhipeng.zhang@nlpr.ia.ac.cn}
\and
Houwen Peng\thanks{corresponding author}\\
Microsoft Research\\
{\tt\small houwen.peng@micrsoft.com}\\
}

\date{\parbox{\linewidth}{\centering%
		\today\endgraf\bigskip
		Coordinator 1 \hspace*{3cm} Coordinator 2\endgraf\medskip
		Dept.\ of Physics \endgraf
		ABC College}}


\maketitle

\thispagestyle{empty}   

\begin{abstract}
    Siamese networks have drawn great attention in visual tracking because of their balanced accuracy and speed. However, the backbone networks used in Siamese trackers are relatively shallow, such as AlexNet~\cite{AlexNet}, which does not fully take advantage of the capability of modern deep neural networks. In this paper, we investigate how to leverage deeper and wider convolutional neural networks to enhance tracking robustness and accuracy. We observe that direct replacement of backbones with existing powerful architectures, such as ResNet~\cite{ResNet} and Inception~\cite{Inception}, does not bring improvements. The main reasons are that 1) large increases in the receptive field of neurons lead to reduced feature discriminability and localization precision; and 2) the network padding for convolutions induces a positional bias in learning. To address these issues, we propose new residual modules to eliminate the negative impact of padding, and further design new architectures using these modules with controlled receptive field size and network stride. 
    The designed architectures are lightweight and guarantee real-time tracking speed when applied to SiamFC~\cite{siamFC} and SiamRPN~\cite{siamRPN}.
    Experiments show that solely due to the proposed network architectures, our SiamFC+ and SiamRPN+ obtain up to 9.8\%/5.7\% (AUC), 23.3\%/8.8\% (EAO) and 24.4\%/25.0\% (EAO) relative improvements over the original versions~\cite{siamFC, siamRPN} on the OTB-15, VOT-16 and VOT-17 datasets, respectively.
\end{abstract}
\vspace{-1.8em}
\section{Introduction}
\vspace{-0.5em}

Visual tracking is one of the fundamental problems in computer vision. It aims to estimate the position of an arbitrary target in a video sequence, given only its location in the initial frame. Tracking at real-time speeds plays a vital role in numerous vision applications, such as surveillance, robotics, and human-computer interaction~\cite{BACF,LiXi_survey,PAMI_survey,zhanglei}.

\begin{figure}[t]
	\vspace{-1.2em}
		\begin{center}
			\includegraphics[width=0.88\linewidth]{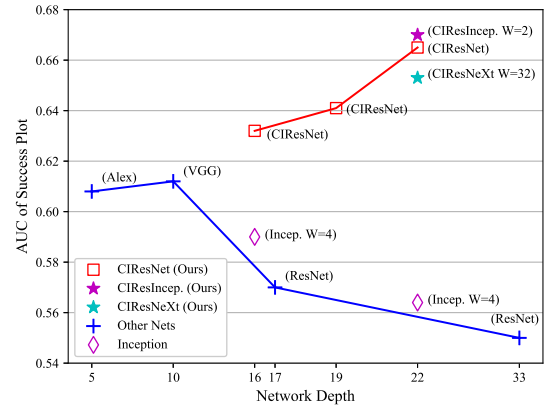}
		\end{center}
	\vspace{-2em}
	\caption{
		AUC of success plot \emph{vs.}~network depth and width (indicated by W). Here, width refers to the number of branches in a module. The results are obtained using SiamFC~\cite{siamFC} with different backbone networks, through evaluation on OTB-13.
	}
	\label{Fig1}
	\vspace{-2.2em}
\end{figure}

Recently, trackers based on Siamese networks~\cite{siamFC,triSiam,DSiam,saSiam,siamRPN,SINT,structSiam} have drawn great attention due to their high speed and accuracy. However, the backbone network utilized in these trackers is still the classical AlexNet~\cite{AlexNet}, rather than modern deep neural networks that have proven more effective for feature embedding.
To examine this issue, we replace the shallow backbone with deeper and wider networks, including \mbox{VGG~\cite{VGG}}, Inception~\cite{Inception} and ResNet~\cite{ResNet}. Unexpectedly, this straightforward replacement does not bring much improvement, and can even cause substantial performance drops when the network depth or width increases, as shown in Fig.~\ref{Fig1}. This phenomenon runs counter to the evidence that increasing network depth and width is beneficial for elevating model capability~\cite{ResNet, Inception}.


One intuitive reasoning is that these deeper and wider network architectures are primarily designed for image classification tasks, where the precise localization of the object is not paramount. To investigate the concrete reason, we analyze the Siamese network architecture and identify that the receptive field size of neurons, network stride and feature padding are three important factors affecting tracking accuracy. In particular, the receptive field determines the image region used in computing a feature. A larger receptive field provides greater image context, while a small one may not capture the structure of target objects. The network stride affects the degree of localization precision, especially for small-sized objects. Meanwhile, it controls the size of output feature maps, which affects feature discriminability and detection accuracy.  Moreover, for a fully-convolutional architecture~\cite{siamFC}, the feature padding for convolutions induces a potential position bias in model training, such that when an object moves near the search range boundary, it is difficult to make an accurate prediction. 
These three factors together prevent Siamese trackers from benefiting from current deeper and more sophisticated network architectures. 

In this paper, we address these issues by designing new residual modules and architectures that allow deeper and wider backbone networks to unleash their power in Siamese trackers. First, we propose a group of cropping-inside residual (CIR) units based on the ``bottleneck'' residual block~\cite{ResNet}. The CIR units crop out padding-affected features inside the block (i.e., features receiving padding signals), and thus prevent convolution filters from learning the position bias. Second, we design two kinds of network architectures, namely deeper and wider networks, by stacking the CIR units. In these networks, the stride and neuron receptive field are formulated to enhance localization precision. 
Finally, we apply the designed backbone networks to two representative Siamese trackers: SiamFC~\cite{siamFC} and SiamRPN~\cite{siamRPN}. Experiments show that solely due to the proposed network architectures, the Siamese trackers obtain up to 9.8\%/5.7\%(AUC), 23.3\%/8.8\%(EAO) and 24.4\%/25.0\%(EAO) relative improvements over the original versions~\cite{siamFC, siamRPN} on the OTB-15, VOT-16 and VOT-17 datasets, respectively.
In addition, the designed architectures are lightweight and permit the trackers to run at real-time speed.

The main cotributions of this work are twofold.
\vspace{-0.6em}
\begin{itemize}[leftmargin=0.55cm]
    \item{
        We present a systematic study on the factors of backbone networks that affect tracking accuracy, and provides architectural design guidelines for the Siamese tracking framework.
    }
 \vspace{-0.5em}
    \item{ We design new deeper and wider network architectures for Siamese trackers, based on our proposed no-padding residual units. Experimental results demonstrate that the new architectures provide clear improvements over the baseline trackers. Code and models are available at~\url{https://github.com/researchmm/SiamDW}.
    }
\vspace{-0.5em}
\end{itemize}

In the remainder of this paper, we first review background on Siamese tracking in Sec.~\ref{Sec2}. This is followed by an analysis of performance degradation in Sec.~\ref{Sec3}. Based on the analysis, we propose new residual modules and network architectures in Sec.~\ref{Sec4}. Experiments and comparisons are reported in Sec.~\ref{Sec5}. We end the paper with a discussion of related work and the conclusion in Sec.~\ref{Sec6} and~\ref{Sec7}.


\vspace{-1em}
\section{Background on Siamese Tracking}
\label{Sec2}

\vspace{-0.5em}

Before analyzing the reasons for the performance degradation shown in Fig.~\ref{Fig1}, we briefly review the fully-convolutional Siamese tracker SiamFC~\cite{siamFC}, which serves as the basic framework discussed in this work. The standard Siamese architecture takes an image pair as input, comprising an exemplar image $\bm{z}$ and a candidate search image $\bm{x}$. The image $\bm{z}$ represents the object of interest (e.g., an image patch centered on the target object in the first video frame), while $\bm{x}$ is typically larger and represents the search area in subsequent video frames. Both inputs are processed by a ConvNet $\varphi$ with parameters $ \theta $. This yields two feature maps, which are cross-correlated as
\vspace{-0.8em}
\begin{equation}
f_\theta(\bm{z}, \bm{x}) = \varphi_\theta(\bm{z})  \star \varphi_\theta(\bm{x}) + b \cdot \bm{\mathbbm{1}},
\vspace{-0.8em}
\label{Eq1}
\end{equation}
where $b \cdot \bm{\mathbbm{1}}$ denotes a bias term which takes the value $b \in \mathbb{R}$ at every location. Eq.~\ref{Eq1} amounts to performing an exhaustive search of the pattern $\bm{z}$ over the image $\bm{x}$. The goal is to match the maximum value in response map $\bm{f}$ to the target location.

To achieve this goal, the network is trained offline with random image pairs $(\bm{z}, \bm{x})$ taken from training videos and the corresponding ground-truth label $\bm{y}$. The parameters $\theta$ of the ConvNet are obtained by minimizing the following logistic loss $\ell$ over the training set:
\vspace{-0.8em}
\begin{equation}
{\arg} \mathop {\min}\limits_{\theta}  \mathop {\mathbb{E}}\limits_{(\bm{z},\bm{x},\bm{y})} \ell(\bm{y}, f_\theta(\bm{z},\bm{x})).
\vspace{-0.7em}
\end{equation}

Previous methods~\cite{siamFC, DSiam, saSiam,siamRPN,structSiam,  DaSiam} commonly utilize the classic and relatively shallow AlexNet~\cite{AlexNet} as the backbone network $\varphi$ in this framework. In our work, we study the problem of how to design and leverage a more advanced ConvNet $\varphi$ to learn an effective model $\theta$ that enhances tracking robustness and accuracy.


\begin{table*}[!t]
	\vspace{-2em}  
    \begin{center}

        \caption{Analysis of network internal factors on AlexNet, VGG-$10$, Incep.-$22$ and ResNet-$33$.  The numbers \ding{172}-\ding{181} represent different versions, in which the convolution kernel size, downsampling layer and padding are modified to show the trends. Details on the modifications are given in the supplementary material due to limited space.} 
        \vspace{-0.9em}
        \fontsize{9pt}{4.5mm}\selectfont
        \begin{threeparttable}
            \begin{tabular}{ @{}c@{} | @{}c@{} @{}c@{} @{}c@{}  @{}c@{} @{}c@{} @{}c@{} @{}c@{} @{}c@{} @{}c@{} @{}c@{} @{}c@{} @{}c@{} | @{}c@{} @{}c@{} @{}c@{} @{}c@{} @{}c@{} @{}c@{} @{}c@{} @{}c@{} @{}c@{}}
                \cline{1-11} \cline{13-22}
                \# NUM & \ding{172} & \ding{173} & \ding{174}  & \ding{175} & \ding{176} & \ding{177} & ~\ding{178} & \ding{179} & \ding{180} & \ding{181} & ~~  & \# NUM   & \ding{172} & \ding{173} & \ding{174}  & \ding{175} & \ding{176}   & \ding{177} & \ding{178} & \ding{179}  & \ding{180}
                \\
                \cline{1-11} \cline{13-22}
                ~~~~~RF\tnote{1}~~~~~ & ~Max(127) & ~+24~~ & ~+16~~ & ~+8~~ & $\pm0$ (87)& ~$\pm0$ ~~& ~~~-8~~~& ~~-16~~ & +16~ & ~~+16~ & & ~~~~~~RF~~~~~~&~~+32~ & ~+16~~ & ~~+8~~ & $\pm0$ (91) & $\pm0$& ~~-8~~ & ~-16~~ & ~+16~ & ~+16~
                \\

                STR & 8 & 8 & 8 & 8 & 8 & 8 & ~8 & 8 & 16 & 4& &STR& ~8 & 8 & 8 & 8 & 8 &8& 8 & 16 & 4
                \\

                OFS & 1 & 3& 4& 5 & 6&16& ~7 & 8 & 2 & 7 & &OFS& ~1 & 3 &4 & 5 & 16 & 6 & 7 & 2 & 6
                \\
                PAD & \ding{55} & \ding{55} & \ding{55} & \ding{55} & \ding{55} & \ding{51} &~\ding{55} & \ding{55} & \ding{55} & \ding{55} & &PAD &~\ding{55} &\ding{55} & \ding{55} & \ding{55} & \ding{51} & \ding{55} & \ding{55} & \ding{55}& \ding{55}

                \\

                \cline{1-11} \cline{13-22}

                \textbf{Alex} & 0.56 & 0.57 &0.60 & ~0.60 & 0.61& 0.55& ~0.59 & 0.58 & 0.55 & 0.59 &   &  \textbf{ResNet}&~ 0.56 &  0.59 & 0.60 & 0.62 & 0.56 &~~0.60~~& 0.60& 0.54 &0.58
                \\\cline{1-11} \cline{13-22}

                \textbf{VGG} & 0.58 & 0.59 &0.61 & ~0.61 & 0.62& 0.56& ~0.59 & 0.58 & 0.54 & 0.58 &  &\textbf{Incep.}\tnote{2} & ~ 0.58 & 0.60 & 0.61 & 0.63 & 0.58 & ~~0.62~~& 0.61& 0.56 &0.59
                \\\cline{1-11} \cline{13-22}
            \end{tabular}

            \begin{tablenotes}
                \footnotesize
                \item[1] To better show the trends, we denote $\pm0$ as the original RF size of the network. $+$ and $-$ represent increasing and decreasing size over the originals. Max($127$) represents the maximum effective RF, which is identical to the size of the exemplar image, i.e. $127\small\times127$ pixels.
                \item[2] For the Inception network, its RF size lies in a range. Here we only list the theoretically maximum size, to align with ResNet for comparison. 
            \end{tablenotes}
        \end{threeparttable}
        \vspace{-0.7em}
        \vspace{-0.5em}
        \label{Tab2}
    \end{center}
\end{table*}

\begin{table}[!t]
    \vspace{-1.3em}
    \begin{center}
        \fontsize{8.5pt}{4mm}\selectfont
        \begin{threeparttable}
            \begin{tabular}{ @{}c@{} | @{}c@{} @{}c@{} @{}c@{} @{}c@{} @{}c@{} @{}c@{} @{}c@{} @{}c@{} @{}c@{} @{}c@{} @{}c@{} @{}c@{} @{}c@{} @{}c@{} @{}c@{} @{}c@{}}
                \cline{1-7}
                ~~~~~~~~~~~~~& ~~Alex~~ & ~~VGG-10~~ & ~~Incep.-16~~ & ~~Res.-17~~ & ~~Incep.-22 & ~~Res.-33~~
                \\ \cline{1-7}
                RF & 87 & 103 & 23\textasciitilde183 & 227 & 39\textasciitilde519 & 739
                \\
                STR & 8 & 8 & 8 & 8 & 16 & 16
                \\
                OFS & 6 & 4 & 16 & 16 & 8 & 8
                \\
                PAD & \ding{55} & \ding{55} & \ding{51} & \ding{51} & \ding{51} & \ding{51}
                \\
                W & 1 & 1 & 4 & 1 & 4 & 1
                \\\cline{1-7}
                \textbf{AUC} & 0.61 & 0.61 &0.59 & 0.57 & 0.56 & 0.55
                \\\cline{1-7}
            \end{tabular}
        \end{threeparttable}
\vspace{-1em}
       	\caption{Internal factors of different networks: receptive field (RF) of neurons in the last layer of network, stride (STR), output feature size (OFS), padding (PAD) and  width (W). Since Inception contains multiple branches in one block, its RF lies within a range.}
        \label{Tab1}
    \end{center}
\vspace{-3em}
\end{table}

\vspace{-0.7em}
\section{Analysis of Performance Degradation}
\label{Sec3}
\vspace{-0.5em}

\label{analysis}
In this section, we analyze the underlying reasons for the performance degradation presented in Fig.~\ref{Fig1}. We conduct ablation experiments on the internal factors of network structures, and identify the ones most responsible for performance drops. We then propose a set of practical guidelines for network architecture design, aimed to alleviate the negative effects.


\vspace{-0.5em}
\subsection{Analysis}
\vspace{-0.5em}
\textbf{Quantitative analysis.} Performance degradation can directly be attributed to network structure, as it is the only setting that changes in the experiments of Fig.~\ref{Fig1}. Therefore, we first identify the structural differences among these network architectures\footnote{Note that the network structures are slightly different from their original versions~\cite{ResNet, VGG, Inception}, where the network stride and padding are modified according to SiamFC~\cite{siamFC}.}. 
As shown in Tab.~\ref{Tab1}, besides depth and width, there are several other internal network factors that differ among the networks, including stride (STR), padding (PAD), receptive field (RF) of neurons in the last layers, and output feature size (OFS). 

To investigate the impact of these factors, we conduct an ablation study. 
We modify the structures of AlexNet, VGG, Inception and ResNet, and expose the effects of the internal factors.
As shown in Tab.~\ref{Tab2}, when network stride (STR) increases from $4$ or $8$ to $16$, the performance drops significantly (\ding{181} \emph{vs}. \ding{174} \emph{vs}. \ding{180} on AlexNet and VGG, \ding{180} \emph{vs}. \ding{173} \emph{vs}. \ding{179} on Incep. and ResNet). This illustrates that Siamese trackers prefer mid-level features (stride $4$ or $8$), which are more precise in object localization than high-level features (stride $\geq 16$). 
For the maximum size of receptive field (RF), the optima lies in a small range.
Specifically, for AlexNet, it ranges from $87$--$8$ (Alex\ding{178}) to $87$+$16$ (Alex\ding{174}) pixels; while for Incep.-$22$, it ranges from $91$--$16$ (Incep.\ding{178}) to $91$+$8$ (Incep.\ding{174}) pixels. VGG-$10$ and ResNet-$17$ also exhibit similar phenomena.
In these cases, the optimal receptive field size is about $60$\%$\sim$$80$\% of the input exemplar image $\bm{z}$ size (e.g. 91 \emph{vs} 127). Intriguingly, this ratio is robust for various networks in our study, and it is insensitive to their structures. It illustrates that the size of RF is crucial for feature embedding in a Siamese framework. The underlying reason is that RF determines the image region used in computing a feature. A large receptive field covers much image context, resulting in the extracted feature being insensitive to the spatial location of target objects. On the contrary, a small one may not capture the structural information of objects, and thus it is less discriminative for matching. Therefore, only RF in a certain size range allows the feature to abstract the characteristics of the object, and its ideal size is closely related to the size of the exemplar image.
For the output feature size, it is observed that a small size (OFS $\leq 3$) does not benefit tracking accuracy. This is because small feature maps lack enough spatial structure description of target objects, and thus are not robust in image similarity calculation.
Moreover, as shown in Tab.~\ref{Tab2} (\ding{176} \emph{vs}. \ding{177} on AlexNet and VGG, \ding{175} \emph{vs}. \ding{176} on Incep. and ResNet), we observe that network padding has a highly negative impact on the final performance. To examine this further, we conduct a qualitative experiment.

\begin{figure}[!t]
    \vspace{-1.3em}  
	\begin{center}
        \includegraphics[width=0.43\textwidth]{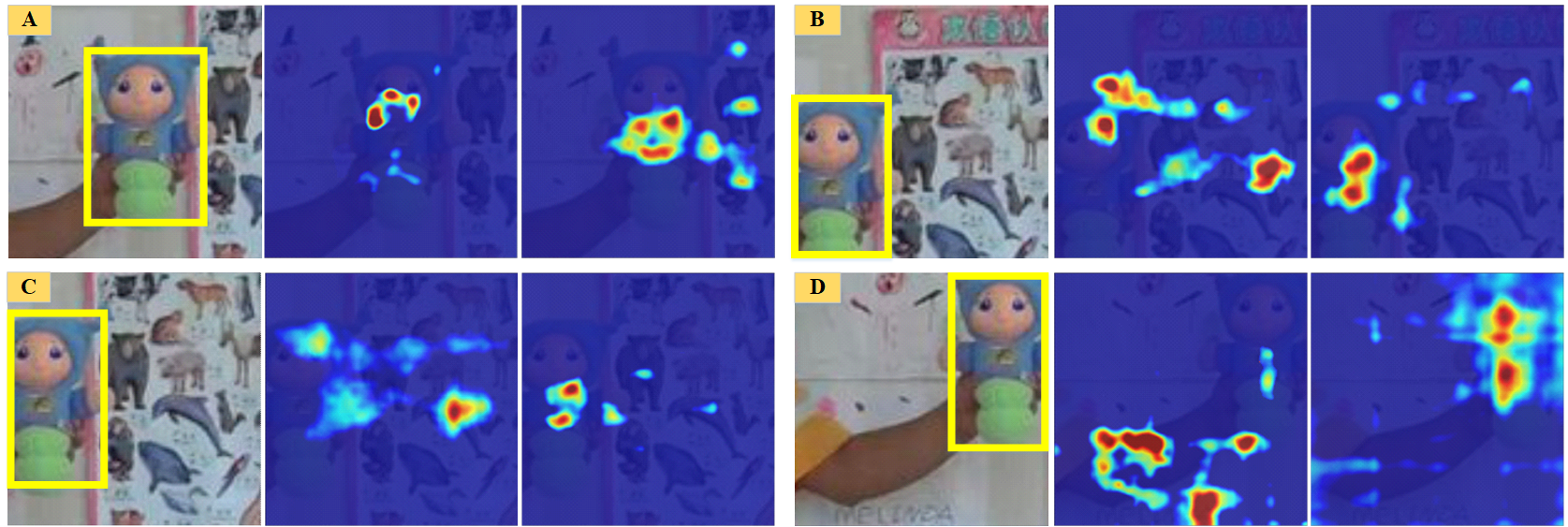}
        \put(-270,  -8){ \scriptsize ~~~~~~~~~~~~~~~~~~~~~~~~~~~~~~ Input ~~~~~~~~~ResNet\ding{176}  ~~CIResNet-22  ~~~~~~Input ~~~~~~~~~ResNet\ding{176}   ~~CIResNet-22}
    \end{center}
    \vspace{-2em}
    \caption{Visualization of position bias learnt in the model w/ and w/o padding (ResNet\ding{176} in Tab.~\ref{Tab1} \emph{vs.} ours). (A) presents the target at the image center, while (B-D) show it moving to boundaries due to imprecise tracked position in the previous frame.}
    \label{Fig2}
    \vspace{-1.5em}   
\end{figure}

\begin{figure*}[!t]
	\vspace{-3em}  
	\centering
	\includegraphics[width=1\linewidth]{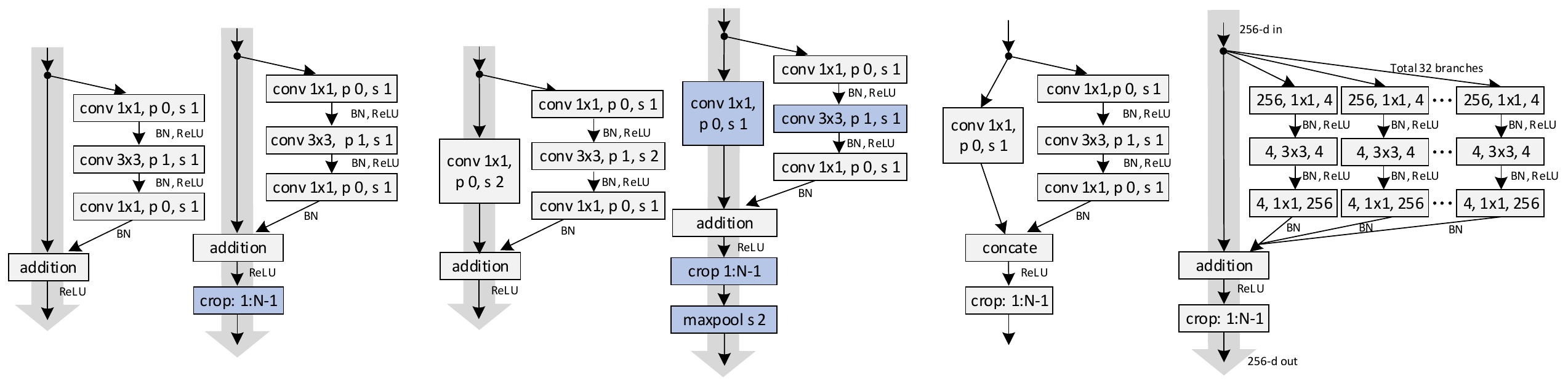}
	\put(-540,  8){ \footnotesize ~~~~~~~~~~~~~~~~~~~~~~~~~~~ ($a$) original ~~~~~~~~~~~~~~~($a^\prime$) CIR  ~~~~~~~~~~~~~~~~~~~ ($b$) original ~~~~~~~~~~~~~~~~~~~~~~~~~~~($b^\prime$) CIR-D   ~~~~~~~~~~~~~~~~ ($c$) CIR-Inception ~~~~~~~~~~~~~~~~~~~($d$) CIR-NeXt}
	\vspace{-1.1em}
	\caption{The proposed cropping-inside residual units. ($a$) and ($b$) are the original residual unit and downsampling unit, while ($a^\prime$) and ($b^\prime$) are our proposed ones. ($c$) and ($d$) are the proposed wide residual units. The grey arrows indicate the shortcut paths for easy information propagation, while the blue boxes highlight the differences from the original units. The letters `p' and `s' indicate the padding size and stride, respectively. The settings of `p' and `s' in ($d$) are the same as in ($c$).}
	\label{units}
	\vspace{-1.7em}
\end{figure*}

\textbf{Qualitative analysis.}  
The Siamese networks feed pairs of exemplar and search images as training data, and learn a feature embedding for matching. If the networks contain padding operations, the embedding features of an exemplar image are extracted from the original exemplar image plus additional (zero)-padding regions. Differently, for the features of a search image, some of them are extracted only from image content itself, while some are extracted from image content plus additional (zero)-padding regions (e.g. the features near the border). 
As a result, there is inconsistency between the embeddings of target object appearing at different positions in search images, and therefore the matching similarity comparison degrades. 
Fig.~\ref{Fig2} presents a visualization example of such inconsistency-induced effect in the testing phase. It shows that when the target object moves to image borders, its peak does not precisely indicate the location of the target. This is a common case caused by tracker drifts, when the predicted target location is not precise enough in the previous frame. %

\vspace{-0.5em}
\subsection{Guidelines}
\vspace{-0.5em}

According to the above analysis, we summarize four basic guidelines to alleviate the negative impacts of structural factors in a network architecture.

\begin{itemize}[leftmargin=0.4cm]
\vspace{-0.6em}
    \item \emph{Siamese trackers prefer a relatively small network stride.} Network stride affects the overlap ratio of receptive fields for two neighboring output features. It thereby determines the basic degree of localization precision. Therefore, when network depth increases, the stride should not increase accordingly. With regards to accuracy and efficiency, an empirically effective choice is to set the stride to 4 or 8.
\vspace{-0.4em}
    \item \emph{The receptive field of output features should be set based on its ratio to the size of the exemplar image.} An empirically effective ratio is $60$\%$\sim$$80$\% for an exemplar image. A proper ratio allows the network to extract a set of features, each of which captures the information of different spatial parts of a target object. This leads the extracted features to be robust in calculating region similarity. Particularly, the maximum RF should not be larger than the exemplar image, otherwise the performance will drop significantly.
\vspace{-0.4em}
    \item \emph{Network stride, receptive field and output feature size should be considered as a whole when designing a network architecture.} These three factors are not independent of one another. If one changes, the others will change accordingly. Considering them together can help the designed network to extract more discriminative features in a Siamese framework.
\vspace{-0.4em}
    \item \emph{For a fully convolutional Siamese matching network, it is critical to handle the problem of perceptual inconsistency between the two network streams.} There are two feasible solutions. One is to remove the padding operations in networks, while the other is to enlarge the size of both input exemplar and search images, and then crop out padding-affected features. 
\vspace{-1em}

\end{itemize}


\begin{table*}[!t]
	\vspace{-2em}  
	\begin{center}
		\caption{Architectures of designed backbone networks for Siamese trackers. The CIR-D units are used in the first block of the `conv3' stage, except for CIResNet-43, in which CIR-D is located in the fourth block.}
		\vspace{-0.8em}
		\fontsize{7.5pt}{3.5mm}\selectfont
		\begin{threeparttable}
			\begin{tabular}{@{}c@{}|@{}c@{}|@{}c@{}|@{}c@{}|@{}c@{}|@{}c@{}|@{}c@{}}
				\cline{1-7}
				~~Stage~~ & ~~CIResNet-16~~ & ~~CIResNet-19~~ & ~~~CIResNet-22~~ & ~~CIResInception-22~~ & ~~CIResNeXt-22~~ & ~~CIResNet-43~~
				\\ \cline{1-7}
				
				conv1 & \multicolumn{6}{@{}c@{}}{~~~7$\times7, 64,$ stride $2$~~~}
				\\ \cline{1-7}
				
				{\multirow{5}{*}{conv2}} &  \multicolumn{5}{c|}{~~~2$\times2$ max pool, stride $2$~~~} &
				\\ \cline{2-6}
				
				& $\begin{bmatrix} 1\times1,  64~~\\ 3\times3,  64~~\\ 1\times1,  256\end{bmatrix}\times 1$ ~~
				&$\begin{bmatrix} 1\times1,  64~~\\ 3\times3,  64~~\\ 1\times1,  256\end{bmatrix}\times 2$ ~~
				&$\begin{bmatrix} 1\times1,  64~~\\ 3\times3,  64~~\\ 1\times1,  256\end{bmatrix}\times 3$ ~~
				&$\begin{bmatrix} 1\times1,  64~~\\ 3\times3,  64~~\\ 1\times1,  256\end{bmatrix}\times 3$ ~~
				&$\begin{bmatrix} 1\times1,  64~~~~~~~~~~~~~~~~~~\\ 3\times3,  64, C=32~~\\ 1\times1,  256~~~~~~~~~~~~~~~~~\end{bmatrix}\times 3$ ~~
				&$\begin{bmatrix} 1\times1,  64~~\\ 3\times3,  64~~\\ 1\times1,  256\end{bmatrix}\times 14$ ~~
				\\
				& & & & $  \hspace{-3mm} \textbf{[} 1\times1, 64 \textbf{]} \times 3$   & &
				\\\cline{1-7}
				
				{\multirow{2}{*}{conv3}}
				&$\begin{bmatrix} 1\times1,  128~~\\ 3\times3,  128~~\\ 1\times1,  512\end{bmatrix}\times 4$ ~~
				&$\begin{bmatrix} 1\times1,  128~~\\ 3\times3,  128~~\\ 1\times1,  512\end{bmatrix}\times 4$ ~~
				&$\begin{bmatrix} 1\times1,  128~~\\ 3\times3,  128~~\\ 1\times1,  512\end{bmatrix}\times 4$ ~~
				&$\begin{bmatrix} 1\times1,  128~~\\ 3\times3,  128~~\\ 1\times1,  512\end{bmatrix}\times 4$ ~~~
				&$\begin{bmatrix} 1\times1,  128~~~~~~~~~~~~~~~~~~\\ 3\times3,  128, C=32~~\\ 1\times1,  512~~~~~~~~~~~~~~~~~\end{bmatrix}\times 4$ ~~
				\\
				& & & & $ \hspace{-3mm} \textbf{[} 1\times1,128 \textbf{]} \times 4$    & &
				\\\cline{1-7}

				&  \multicolumn{6}{@{}c@{}}{cross correlation Eq.~\ref{Eq1}}
				\\\cline{1-7}
				
				\# RF & 77 & 85 & 93 & 13$\sim$93 & 93 & 105
				\\\cline{1-7}
				\# OFS & 7& 6& 5& 5& 5 &6
				\\\cline{1-7}
				\# Params &1.304 M & 1.374 M& 1.445 M& 1.695 M& 1.417 M& 1.010 M
				\\\cline{1-7}
				\# FLOPs & 2.43 G& 2.55 G& 2.65 G& 2.71 G & 2.52 G & 6.07 G
				\\\cline{1-7}
				
			\end{tabular}
		\end{threeparttable}
		\label{arch}
	\end{center}
	\vspace{-2.5em}
\end{table*}

\vspace{-0.5em}
\section{Deeper and Wider Siamese Networks}
\label{Sec4}
\vspace{-0.6em} 

In this section, we design new modules, i.e. cropping-inside residual (CIR) units, to eliminate the underlying position bias.  Then, we build up deeper and wider backbone networks by stacking the new residual modules. The network stride and receptive field size are well controlled according to the guidelines. We further apply the designed networks to two representative Siamese trackers, i.e. SiamFC~\cite{siamFC} and SiamRPN~\cite{siamRPN}.

\vspace{-0.9em}
\subsection{Cropping-Inside Residual (CIR) Units}
\vspace{-0.3em}

The residual unit~\cite{ResNet} is a key module in network architecture design due to its easy optimization and powerful representation. It consists of $3$ stacked convolution layers and a shortcut connection that bypasses them, as shown in Fig.~\ref{units}($a$). The three layers are $1\small{\times}1$, $3 \small{\times} 3$ and $1\small{\times}1$ convolutions, where the $1\small{\times}1$ layers are responsible for reducing and then restoring dimensions, leaving the $3\small{\times}3$ layer as a bottleneck with smaller input and output dimensions. This bottleneck convolution includes zero-padding of size $1$, to ensure a compatible output size before the addition.

\textbf{CIR Unit.}
As discussed in Sec.~\ref{analysis}, network padding may introduce position bias in the Siamese framework. Hence, it is necessary to remove the padding in residual units when utilizing them to build a Siamese network. To this end, we augment the residual unit with a cropping operation, which is incorporated after the feature addition, as shown in Fig.~\ref{units}($a^\prime$). The cropping operation removes features whose calculation is affected by the zero-padding signals. Since the padding size is one in the bottleneck layer, only the outermost features on the border of the feature maps are cropped out. This simple operation neatly removes padding-affected features in residual unit.

\textbf{Downsampling CIR (CIR-D) Unit.} The downsampling residual unit is another key building block for network design. It is utilized to reduce the spatial size of feature maps while doubling the number of feature channels. Similar to the residual unit, the downsampling unit also contains padding operations, as shown in Fig.~\ref{units}($b$). Thus, we also modify its structure to remove the negative effects caused by the padding. As shown in Fig.~\ref{units}($b^\prime$), we change the convolutional stride from $2$ to $1$ within both the bottleneck layer and shortcut connection. Cropping is again inserted after the addition operation to remove the padding-affected features. Finally, max-pooling is employed to perform spatial downsampling of the feature map. The key idea of these modifications is to ensure that only the features influenced by padding are removed, while keeping the intrinsic block structure unchanged. If we were only to insert cropping after the addition operation, as done in the proposed CIR unit, without changing the position of downsampling, the features after cropping would not receive any signal from the outermost pixels in the input image. As the network depth increases, this would effectively cause even more image content to be removed, resulting in noisy/incomplete extracted features.

\textbf{CIR-Inception and CIR-NeXt Units.}
We further equip the CIR unit with a multi-branch structure, enabling it to be used in building wide networks. Similar to Inception~\cite{Inception} and ResNeXt~\cite{ResNeXt}, we widen the CIR unit with multiple feature transformations, generating the CIR-Inception and CIR-NeXt modules as shown in Fig.~\ref{units}($c$-$d$). Specifically, in the CIR-Inception structure, we insert a $1\small{\times}1$ convolution into the shortcut connection, and merge the features of the two branches by concatenation, rather than by addition. In CIR-ResNeXt, we split the bottleneck layer into $32$ transformation branches, and aggregate them by addition. Moreover, for the downsampling units of CIR-Inception and CIR-NeXt, the modifications are the same as those in CIR-D (Fig.~\ref{units}($b^\prime$)), where the convolution stride is reduced and max-pooling is added. These two multi-branch structures enable the units to learn richer feature representations.



\vspace{-0.8em}
\subsection{Network Architectures}
\vspace{-0.45em}

By stacking the above CIR units, we build up deeper and wider networks. The constructions follow our design guidelines. First, we determine the network stride. A stride of $8$ is used to build a $3$-stage network, while a stride of $4$ is employed in constructing a $2$-stage one. Then, we stack CIR units. We control the number of units and the position of downsampling units in each stage. The goal is to ensure that the receptive field size of neurons in the final layer lies within the derived range, i.e. $60$\%-$80$\% of the exemplar image. Additionally, when network depth increases, the receptive field may exceed this range. Therefore, we halve the stride to $4$ to control the receptive field.

\textbf{Deeper Networks.} We construct deeper networks using CIR and CIR-D units. The structures are similar to ResNet~\cite{ResNet}, but with different network stride, receptive field size, and building blocks. In Tab.~\ref{arch}, we present four deep cropping-inside residual networks, i.e. CIResNet-$16$, $19$, $22$ and $43$. Since these networks have similar structure, we present details for just two of them: CIResNet-$22$ and CIResNet-$43$.

CIResNet-$22$ has $3$ stages (stride=$8$) and consists of $22$ weighted convolution layers. Except for the first $7\small{\times}7$ convolution, the others are all CIR units. A cropping operation (with size of $2$) follows the $7\small{\times}7$ convolution to remove padding-affected features. The feature downsampling in the first two stages are performed by a convolution and a max-pooling of stride $2$, following the original ResNet~\cite{ResNet}. In the third stage, downsampling is performed by the proposed CIR-D unit, which is located at the first block in this stage ($4$ blocks in total). When the feature map size is downsampled, the number of filters is doubled to increase feature discriminability. The spatial size of the output feature map is $5\small{\times}5$, with each feature receiving signals from a region of size $93\small{\times}93$ pixels on the input image plane, i.e. the corresponding size of receptive field.

We further increase network depth to $43$ layers in building CIResNet-$43$. Because of its large depth, CIResNet-$43$ is designed with only $2$ stages, to keep its receptive field size within the suggested range. In the second stage of CIResNet-$43$, there are $14$ blocks, where the fourth one has a CIR-D unit for feature downsampling. It is worth noticing that CIResNet-$43$ almost reaches the maximum depth of backbone networks that can achieve real-time speed in the SiamFC~\cite{siamFC} framework. It has 6.07G FLOPs (multiply-adds) and runs at an average of $\sim$35 fps in SiamFC framework on a GeForce GTX 1080 GPU.

\textbf{Wider Networks.} We construct two types of wide network architectures using CIR-Inception and CIR-NeXt units, respectively. Here, we only present a $22$-layer structure as an example, since other wider networks are similar to this case.
As presented in Tab.~\ref{arch}, the wide networks, i.e. CIResInception-$22$ and CIResNeXt-$22$, have similar structure to CIResNet-22 in terms of network stride, building block number and output feature size. But the network widths are increased by $2$ and $32$ times respectively, through the multi-branch building blocks. Also, the receptive field size becomes diverse (i.e. $13$$\sim$$93$) in CIResInception-$22$ due to multi-branch concatenation, but the maximum size still remain within the suggested range.

\vspace{-0.5em}
\subsection{Applications}
\vspace{-0.4em}

We apply the designed deeper and wider networks to two representative Siamese trackers: the classical SiamFC~\cite{siamFC} and the most recently proposed SiamRPN~\cite{siamRPN}. In both of these two trackers, we replace the original shallow backbones, i.e. the $5$-layer AlexNet~\cite{AlexNet}, with our designed networks, which is the only modification to the original frameworks.

\vspace{-0.6em}
\section{Experiments}
\label{Sec5}
\vspace{-0.6em}

This section presents the results of our deeper and wider Siamese networks on multiple benchmark datasets, with comparisons to the state-of-the-art tracking algorithms. Ablation studies are also provided to analyze the effects of the components in the proposed networks.

\vspace{-0.5em}
\subsection{Experimental Details}
\vspace{-0.4em}

\textbf{Training.} The parameters of our networks are initialized with the weights pre-trained on ImageNet~\cite{ImageNet}. During training, we freeze the weights of the first $7\small{\times}7$ convolution layer, and gradually fine-tune other layers from back to front. We unfreeze the weights of the layers in each block (i.e. the proposed cropping-inside residual units) after every five training epochs. There are $50$ epochs in total, the same as in~\cite{siamFC, siamRPN}. The learning rates are decreased logarithmically from $10^{-3}$/$10^{-2}$ to $10^{-7}$/$10^{-5}$ for SiamFC and SiamRPN, respectively. The weight decay is set to $10^{-4}$, and the momentum is set to $0.9$ (for both SiamFC and SiamRPN). We use synchronized SGD~\cite{SGD} on 4 GPUs, with each GPU hosting 32 images, hence the mini-batch size is 128 images per iteration. 

The training image pairs for SiamFC are collected from the ImageNet VID dataset~\cite{ImageNet}, while for SiamRPN, it is generated from ImageNet VID~\cite{ImageNet} and Youtube-BB~\cite{YTB}, which is the same as those in the original frameworks~\cite{siamFC, siamRPN}. The size of an exemplar image is $127\small{\times}127$ pixels, while the size of a search image is $255\small{\times}255$ pixels.

\begin{table}[!t]
	\centering
	\fontsize{8pt}{4mm}\selectfont

	\begin{threeparttable}
		\begin{tabular}{ @{}r@{} @{}c@{} @{}c@{} @{}c@{} @{}c@{} @{}c@{} @{}c@{} }
			\toprule
			\multirow{2}{*}{Backbone} & \multicolumn{2}{@{}c@{}}{\textbf{OTB}(AUC)} & \multicolumn{2}{@{}c @{}}{\textbf{VOT-17}(EAO)} & \multicolumn{2}{@{}c@{}}{\textbf{FPS}}\cr
			\cmidrule(lr){2-3} \cmidrule(lr){4-5} \cmidrule(lr){6-7}
			& \scriptsize{SiamFC} & \scriptsize{SiamRPN} & ~\scriptsize{SiamFC} & \scriptsize{SiamRPN} & ~\scriptsize{SiamFC}~ & \scriptsize{SiamRPN}
			\\ \cline{1-7}
			AlexNet~ & ~0.608\cite{siamFC}~ & ~0.637\cite{siamRPN}~ & ~0.188\cite{VOT17}~& ~0.244\cite{siamRPN}~~ & 101\cite{siamFC} & 190\cite{siamRPN}
			\\
			CIResNet-16~ & 0.632& 0.651& 0.202& 0.260 & 75 & 160
			\\
			CIResNet-19~ & 0.640& 0.660& 0.225& 0.279 & 73& 155
			\\
			CIResNet-22~ & 0.662& 0.662& \textbf{0.234}& \textbf{0.301} & 70 & 150
			\\
			CIResIncep.-22~ & \textbf{0.666}& \textbf{0.673}& 0.215& 0.296 & 67& 145
			\\
			CIResNeXt-22~ & 0.654& 0.660& 0.230& 0.285 & 72& 155
			\\
			CIResNet-43~ & 0.638& 0.652& 0.207& 0.265 & 35& 75
			\\
			\bottomrule
		\end{tabular}
\vspace{-0.5em}
	\end{threeparttable}
\caption{Performance of our network architectures in SiamFC and SiamRPN. To compare with the original results reported in~\cite{siamFC, VOT17, siamRPN}, SiamFCs are evaluated on OTB-2013 and VOT-17, while SiamRPNs are evaluated on OTB-2015 and VOT-17. The speed (FPS) is measured on a GeForce GTX 1080 GPU.}

	\vspace{-1em}
	\label{baseline}
\end{table}

\begin{table*}[!t]
	\vspace{-2em}   
	\begin{center}
		\fontsize{8.5}{8}\selectfont
		\caption{Performance comparisons on five tracking benchmarks. \textcolor{red}{Red}, \textcolor{green}{Green} and \textcolor{blue}{Blue} fonts indicate the top-3 trackers, respectively. 
		}
		\label{bigtab}
		\vspace{-1.3em}
		\begin{tabular}{llccccccccccccc}
			
			\toprule
			\multirow{2}{*}{Tracker}&\multirow{2}{*}{Year}&
			\multicolumn{2}{c}{OTB-2013}&\multicolumn{2}{c}{OTB-2015}&\multicolumn{3}{c}{VOT15}&\multicolumn{3}{c}{VOT16}&\multicolumn{3}{c}{VOT17}\cr
			\cmidrule(lr){3-4} \cmidrule(lr){5-6} \cmidrule(lr){7-9} \cmidrule(lr){10-12} \cmidrule(lr){13-15}
			& &AUC&Prec.&AUC&Prec.&A&R&EAO&A&R&EAO&A&R&EAO\cr
			\midrule
			SRDCF~\cite{SRDCF}&2015&0.63&0.84&0.60&0.80&0.56&1.24&0.29&0.54&0.42&0.25&0.49&0.97&0.12 \cr
			SINT~\cite{SINT}&2016&0.64&0.85&-&-&-&-&-&-&-&-\cr
			Staple~\cite{Staple}&2016&0.60&0.80&0.58&0.78&0.57&1.39&0.30&0.54&0.38&0.30&0.52&0.69&0.17 \cr
			SiamFC~\cite{siamFC}&2016&0.61&0.81&0.58&0.77&0.53&0.88&0.29&0.53&0.46&0.24&0.50&0.59&0.19\cr
			ECO-HC~\cite{ECO}&2017&0.65&0.87&\textbf{\textcolor{green}{0.64}}&0.86&-&-&-&0.54&0.3&\textbf{\textcolor{blue}{0.32}}&0.49&0.44&\textbf{\textcolor{blue}{0.24}}\cr
			PTAV~\cite{PTAV}&2017&\textbf{\textcolor{blue}{0.66}}&0.89&\textbf{\textcolor{green}{0.64}}&0.85&-&-&-&-&-&-&-&-&-\cr
			DSiam~\cite{DSiam}&2017&0.64&0.81&-&-&-&-&-&-&-&-&-&-&-\cr
			CFNet~\cite{CFNet}&2017&0.61&0.80&0.59&0.78&-&-&-&-&-&-&-&-&-\cr
			StructSiam~\cite{structSiam}&2018&0.64&0.88&0.62&0.85&-&-&-&-&-&0.26&-&-&-\cr
			TriSiam~\cite{triSiam}&2018&0.62&0.82&0.59&0.78&-&-&-&-&-&-&-&-&0.20\cr
			SiamRPN~\cite{siamRPN}&2018&-&-&\textbf{\textcolor{green}{0.64}}&0.85&0.58&1.13&\textbf{\textcolor{green}{0.35}}&0.56&0.26&\textbf{\textcolor{green}{0.34}}&0.49&0.46&\textbf{\textcolor{green}{0.24}}\cr
			\cmidrule(lr){1-15}
			SiamFC+&Ours&\textbf{\textcolor{red}{0.67}}&0.88&\textbf{\textcolor{green}{0.64}}&0.85&0.57&-&\textbf{\textcolor{blue}{0.31}}&0.54&0.38&0.30&0.50& 0.49&0.23\cr
			SiamRPN+&Ours&\textbf{\textcolor{red}{0.67}}&0.92&\textbf{\textcolor{red}{0.67}}&0.90&0.59&-&\textbf{\textcolor{red}{0.38}}&0.58&0.24&\textbf{\textcolor{red}{0.37}}&0.52&0.41&\textbf{\textcolor{red}{0.30}}\cr
			\bottomrule
		\end{tabular}
		\vspace{-2.5em}
	\end{center}
\end{table*}

\textbf{Testing.} Tracking follows the same protocols as in SiamFC~\cite{siamFC} and SiamRPN~\cite{siamRPN}. The embedding $\varphi_\theta(\bm{z})$ of the target object is computed once at the first frame, and then is continuously matched to subsequent search images $\varphi_\theta(\bm{x})$.
To handle scale variations, SiamFC searches for the object over three scales $1.0482^{\{-1,0,1\}}$ and updates the scale by linear interpolation with a factor of $0.3629$ to provide damping. SiamRPN searches over only one scale since it employs proposal refinement to handle scale change. The penalty for large change of proposal size and aspect ratio is set to $0.439$.

Our networks and trackers are implemented using Python $3.6$ and PyTorch $0.3.1$. The experiments are conducted on a PC with a GeForce GTX 1080 GPU and a Xeon E5 2.4GHz CPU.

\begin{figure}[!t]
    \caption{Expected average overlap (EAO) plot for VOT-15, 16 and 17. The listed methods, such as EBT\cite{EBT}, LDP\cite{LDP}, nSAMF\cite{SAMF}, TCNN\cite{TCNN}, MLDF\cite{MLDF}, CFWCR\cite{MLDF} and CFCF\cite{CFCF} are compared in VOT challenges~\cite{VOT15,VOT16,VOT17}.}
    \vspace{-1.5em}
    \begin{center}
        \includegraphics[width=0.47\textwidth]{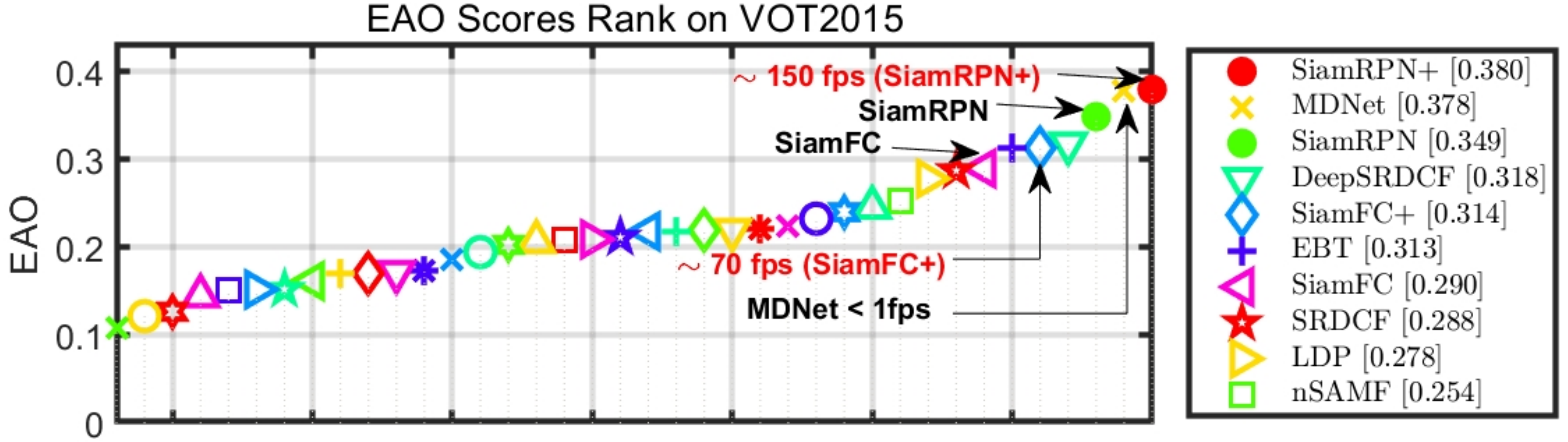}
        \includegraphics[width=0.47\textwidth]{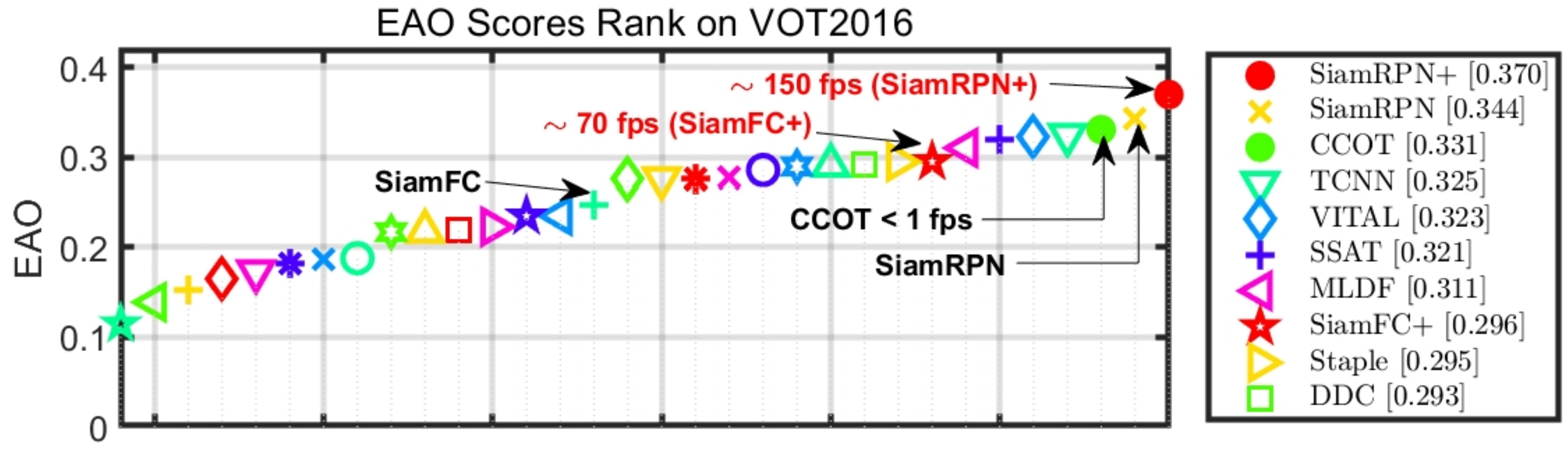}
        \includegraphics[width=0.47\textwidth]{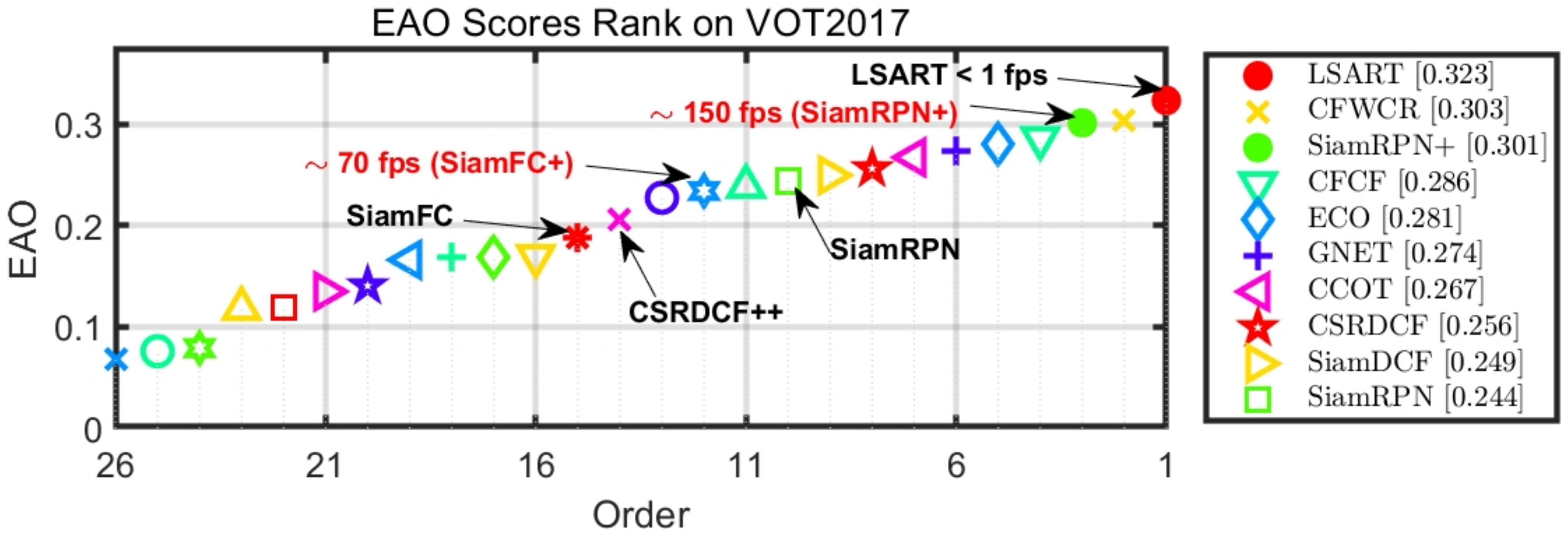}
    \end{center}

    \vspace{-2.5em}

\label{vot}
\end{figure}

\vspace{-0.5em}
\subsection{Comparison with Baselines}
\vspace{-0.4em}
We first compare our deeper and wider networks to the baseline AlexNet in the SiamFC and SiamRPN frameworks. As presented in Tab.~\ref{baseline}, on OTB-13, OTB-15 and VOT-17 datasets, our proposed networks outperform the baseline AlexNet. In particular, SiamFC equipped with a CIResNet.-$22$ backbone obtains relative improvements of $9.8$\% (AUC) and $24.4$\% (EAO) over the original AlexNet on OTB-2015 and VOT-17, respectively. Meanwhile, SiamRPN armed with CIResNet-$22$ achieves $4.4$\% and $23.3$\% relative gains. This verifies that our designed architectures resolve the performance degradation problem shown in Fig.~\ref{Fig1}. Also, it shows the effectiveness of our proposed CIR units for Siamese networks.

It is worth noting that when the depth of CIResNets increases from $16$ to $22$ layers, the performance improves accordingly. But when increasing to $43$ layers, CIResNet does not obtain further gains. There are two main reasons. 1) The network stride is changed to $4$, such that the overlap between the receptive fields of two adjacent features is large.
Consequently, it is not as precise as networks with stride of $8$ in object localization. 2) The number of output feature channels is halved, compared to the other networks in Tab.~\ref{arch} (i.e. $256$ \emph{vs.} $512$ channels). The overall parameter size is also smaller. These two reasons together limit the performance of CIResNet-$43$. Furthermore, wider networks also bring gains for Siamese trackers. Though CIResNeXt-$22$ contain more transformation branches, its model size is smaller (see Tab.~\ref{arch}). Therefore, its performance is inferior to CIResIncep.-$22$ and CIResNet-$22$.

\vspace{-1.2em}
\subsection{Comparison to State-of-the-Art Trackers}
\vspace{-0.4em}

We further compare our enhanced Siamese trackers to state-of-the-art tracking algorithms. We select some of the currently best performing trackers in general, as well as other recent Siamese trackers for comparison. Our enhanced trackers employ the best performing CIResNet-$22$ as the backbone, and are denoted as SiamFC+ and SiamRPN+. The comparison is conducted on five datasets: OTB-2013, OTB-2015, VOT15, VOT16 and VOT17.

\textbf{OTB Benchmarks.} The evaluation on OTB-2013 and OTB-2015 follows
the standard protocols proposed in~\cite{OTB-13, OTB-15}. Two metrics, i.e. precision and area under curve (AUC) of success plots, are used to rank the trackers. The results are reported within Tab.~\ref{bigtab}. It shows that our SiamFC+ and SiamRPN+ surpass other Siamese trackers, such as the recent proposed StructSiam~\cite{structSiam} and TriSiam~\cite{triSiam}. This demonstrates the effectiveness of our designed architecture. Moreover, compared with other state-of-the-art algorithms, such as ECO-HC~\cite{ECO} and CFNet~\cite{CFNet}, our trackers are still superior in terms of precision and speed.



\textbf{VOT Benchmarks.} The evaluation on VOT benchmarks is performed by the official toolkit~\cite{VOT17}, in which accuracy (A), robustness (R) and expected average overlap (EAO) serve as the metrics. 

\emph{VOT-15.} We compare our SiamFC+ and SiamRPN+ with the state-of-the-art trackers on the \emph{vot-2015} challenge. The results are reported in Fig.~\ref{vot} (top). Our SiamRPN+ achieves the best result, slightly better than MDNet~\cite{MDNet}. Also, SiamRPN+ runs much faster than MDNet, which operates at $\sim$$1$ \emph{fps}.
Compared to the baselines, i.e. SiamFC and SiamRPN, our deeper network enhanced trackers obtain $8.8$\% and $8.9$\% relative improvements, respectively.

\emph{VOT-16.} The video sequences in VOT-16 are the same as those in VOT-15, but the ground-truth bounding boxes are precisely re-annotated. We compare our trackers to the top-$10$ trackers in the challenge. As shown in Fig.~\ref{vot}(middle), SiamRPN+ ranks first in terms of EAO. It surpasses CCOT~\cite{CCOT}, the champion of the \emph{vot-2016} challenge, by $3.9$ points, as well as the most recent VITAL~\cite{VITAL} by $4.7$ points.
Moreover, SiamFC+ also improves over the baseline by a large margin, i.e. $6.0$ points on EAO.

\emph{VOT-17.} Fig.~\ref{vot}(bottom) shows the comparison with the trackers in the \emph{vot-2017} challenge. Our SiamRPN+ achieves an EAO of $3.01$, slightly inferior to the best performing LSART tracker~\cite{LSART}.
However, SiamRPN+ runs at $150$ \emph{fps}, which is $150$$\times$ faster than LSART.
Compared to real-time trackers, SiamRPN+ ranks first in terms of accuracy and robustness. Surprisingly, even the plain SiamFC+ surpasses the real-time tracker champion of \emph{vot-2017} CSRDCF++\cite{CSRDCF} by $2.2$ points. This further verifies the effectiveness of our deeper network architecture designed for Siamese trackers.

\begin{table}[!t]
	\centering
	\caption{Ablation for residual unit \emph{vs.} CIR unit on SiamFC. }
	\vspace{-1.1em}
	\fontsize{7.5pt}{4mm}\selectfont
	\begin{threeparttable}
		\begin{tabular}{ @{}r@{} | @{}c@{} @{}c@{} @{}c@{} }
			
			~~~~~~& ~~~CIResNet-$20$~~~ & ~~~CIResNet-$22$~~~ & CIResIncep.-$22$
			\\  \cline{1-4}
			Res. Unit ~~ & ~~0.204 & 0.213 & 0.227
			\\
			\textbf{CIR Unit} ~~ & ~~\textbf{0.271} & \textbf{0.301} & \textbf{0.282}
		\end{tabular}
		
	\end{threeparttable}
	\vspace{-1.7em}  
	
	\label{ablation1}
\end{table}

\begin{table}[!t]
	\centering
	\vspace{-0.3em}
	\caption{Ablation over different downsampling settings used in SiamFC. See main text for explanations.}
	\vspace{-1.1em}
	\fontsize{7.5pt}{4mm}\selectfont
	\begin{threeparttable}
		\begin{tabular}{ @{}r@{} | @{}c@{} @{}c@{} @{}c@{} }
			
			~~~~~~& ~~~CIResNet-$20$~~~ & ~~~CIResNet-$22$~~~ & CIResIncep.-$22$
			\\  \cline{1-4}
			Setting 1 ~~ & ~~0.264 & 0.292 & 0.266
			\\
			Setting 2 ~~ & ~~0.259 & 0.287 & 0.275
			\\
			\textbf{CIR-D} ~~ & ~~\textbf{0.271} & \textbf{0.301} & \textbf{0.282}
		\end{tabular}
	\end{threeparttable}
	\vspace{-3em}
	\label{ablation2}
\end{table}

\vspace{-0.55em}
\subsection{Ablation Study} 
\vspace{-0.45em}

In Tab.~\ref{ablation1}$-$\ref{ablation3}, we evaluate the effects of different factors in our designed networks on the VOT-$16$ dataset.

\emph{With} vs.~\emph{without CIR unit}. The cropping-inside residual unit is a key component in our network architectures. To evaluate its impact, we replace it with the original residual unit~\cite{ResNet} in the networks. As shown in Tab.~\ref{ablation1}, this replacement causes remarkable performance drops, e.g. a degradation of $8.8$ points from $0.301$ to $0.213$ on CIResNet.-$22$. It clearly verifies the importance of padding removal in the CIR unit, which essentially eliminates position bias in learning. The predicted heatmaps of CIResNet-$22$ in Fig.~\ref{Fig2} also prove this point.

\emph{With} vs.~\emph{without CIR-D unit}. 
We compare three different downsampling settings in networks:
1) directly using the original downsampling residual unit, i.e.~Fig.~\ref{arch}($b$), 2) inserting a cropping operation after addition in the downsampling residual unit, and 3) the proposed CIR-D unit, i.e.~Fig.~\ref{arch}($b^\prime$). Tab.~\ref{ablation2} presents the results.  It shows that the first two settings are comparable, but inferior to the third one. This indicates our CIR-D unit is effective. In particular, the cropping introduced in the second setting does not bring improvements, since it removes parts of internal features (i.e. not the padding-affected features), resulting in information loss from the original input.

\emph{Impact of receptive field, feature size and stride.} 
We tune the sizes of these factors and show their impacts on final performance. Specifically, we vary the convolutional kernel size in the last cropping-inside residual block to change the size of receptive field and output feature. Taking CIResNet-$22$ as an example, we vary the kernel size from $1$ to $6$, which causes the feature size to change from $7$ to $2$. To change the network stride, we replace one CIR unit with a CIR-D unit in the networks.
Tab.~\ref{ablation3} shows the results. We can observe that when RF becomes large, the performance drops significantly (i.e. from \ding{175} to \ding{172} in Tab.~\ref{ablation3}). The underlying reason is that a large RF covers much image context, resulting in the extracted feature being insensitive to the spatial location of the target object. For output features, it is observed that a small size (OFS $\leq 3$) does not benefit accuracy. Also, a large network stride, i.e. $16$, is not as precise as a medium sized one, i.e. $8$. These results echo our analysis and guidelines presented at the beginning of this paper.

\begin{table}[!t]

    \begin{center}
        \caption{Analysis of network internal factors.
        }
\vspace{-0.5em}
        \fontsize{7.5pt}{4mm}\selectfont
        \begin{threeparttable}
            \begin{tabular}{ @{}c@{} | @{}c@{} @{}c@{} @{}c@{} @{}c@{} @{}c@{} @{}c@{} @{}c@{} @{}c@{} @{}c@{} @{}c@{} @{}c@{} @{}c@{} @{}c@{} @{}c@{} @{}c@{} @{}c@{} @{}c@{} @{}c@{} @{}c@{}}
                \cline{1-16}
                \# NUM & \ding{172} & \ding{173} & \ding{174}  & \ding{175} & \ding{176} & \ding{177} & \ding{178} & \ding{179} \\
                \cline{1-16}

                RF&~~ +24 ~~ & ~~+16~~ &~~~+8~~  & $\pm0$ (93) & ~~-8 ~~~~& ~~-16~~ & ~~+16~~& ~~+16~~
                \\

                OFS & 2 & 3& 4& 5 &  6 & 7 & 6 & 2
                \\

                STR & 8 &  8 & 8 & 8 &8 & 8& 4 & 16
                \\

                \hline
                \hline

                \textbf{CIResNet-16}~~~~~~~~& 0.22 & 0.23 &0.24 & 0.26 & 0.25 &0.24  &0.23 & 0.20
                \\\cline{1-16}
                \textbf{CIResNet-19}~~~~~~~~& 0.23 & 0.26 &0.26 & 0.28 & 0.27& 0.26  & 0.24& 0.21
                \\\cline{1-16}

                \textbf{CIResNet-22}~~~~~~~~ & 0.25 & 0.27 &0.28 & 0.30 & 0.29& 0.27  & 0.26& 0.23
                \\\cline{1-16}
                \textbf{CIResIncep.-22}~~~ & 0.24 & 0.26 &0.27 & 0.28 & 0.27& 0.26  & 0.25& 0.22
                \\\cline{1-16}
            \end{tabular}

        \end{threeparttable}
\vspace{-3em}
\label{ablation3}
    \end{center}
\end{table}

\vspace{-1.em}
\section{Discussions}
\label{Sec6}
\vspace{-0.7em}

\textbf{Network Architectures.} The problem addressed in this paper can be seen as a subtask of network architecture design, which mainly develops in two ways: making networks deeper~\cite{VGG, ResNet} or wider~\cite{Inception,ResNeXt}. To make networks deeper, ResNets~\cite{ResNet} introduce an identity mapping that makes training ultra deep networks possible. To make networks wider, GoogLeNet~\cite{Inception} and its descendants employ an Inception module to introduce multiple feature transformations and thus enhance model representation capacity.

Our work takes advantage of these deep and wide network architectures, and modifies them to adapt effectively to Siamese networks. Two key principles for Siamese backbone design are presented. One is to remove padding operations inside network architectures, the other is to control the receptive field size and network stride. Both of them are shown to have significant impact on
tracking performance. Moreover, this is the first work that systematically studies how to design robust backbone networks in visual tracking.

\textbf{Siamese Trackers.}
Siamese trackers follow a tracking by similarity matching strategy. The pioneering works are SINT~\cite{SINT} and SiamFC~\cite{siamFC}, which each employ a Siamese network to offline train a similarity metric between the object target and candidate image patches. A large amount of follow-up work~\cite{triSiam, DSiam, saSiam, siamRPN, structSiam} have been proposed, and they fall into two camps. One improves matching precision with high-level semantic information or a localization model~\cite{saSiam, GOTURN, structSiam, siamRPN}. The other enhances the offline Siamese model with online updates~\cite{DSiam, CFNet, DaSiam}.


There is a recent work also studying how to leverage deep networks for visual tracking~\cite{Unveiling}. But it approaches the problem from the direction of data augmentation and feature fusion. Differently, our work studies how to design network architectures and successfully equips Siamese trackers with deeper and wider backbones.

\vspace{-0.7em}
\section{Conclusion}
\label{Sec7}
\vspace{-0.7em}


In this paper, we design deep and wide network architectures for Siamese trackers. This is motivated by the observation that direct replacement of backbones with existing powerful networks does not bring improvements. We carefully study the key causes and determine that receptive field size, network padding and stride are crucial factors. Experiments on five benchmarks demonstrate the effectiveness of the proposed architectures, leading to competitive performance on five datasets.

\noindent
\textbf{Acknowledgement.} This work was done in Microsoft Research Asia. Thanks to Steve Lin and Zhirong Wu for helpful discussions. Zhipeng Zhang is partly supported by NSFC U1803119, 2016QY01W0106 and JQ18018.

{\small
\bibliographystyle{ieee}
\bibliography{deeper_wider_siamese}

\begin{thebibliography}{10}\itemsep=-1pt

\bibitem{Staple}
L.~Bertinetto, J.~Valmadre, S.~Golodetz, O.~Miksik, and P.~H. Torr.
\newblock Staple: Complementary learners for real-time tracking.
\newblock In {\em Proceedings of the IEEE conference on computer vision and
  pattern recognition}, pages 1401--1409, 2016.

\bibitem{siamFC}
L.~Bertinetto, J.~Valmadre, J.~F. Henriques, A.~Vedaldi, and P.~H. Torr.
\newblock Fully-convolutional siamese networks for object tracking.
\newblock In {\em European conference on computer vision}, pages 850--865.
  Springer, 2016.

\bibitem{Unveiling}
G.~Bhat, J.~Johnander, M.~Danelljan, F.~S. Khan, and M.~Felsberg.
\newblock Unveiling the power of deep tracking.
\newblock {\em arXiv preprint arXiv:1804.06833}, 2018.

\bibitem{ECO}
M.~Danelljan, G.~Bhat, F.~S. Khan, M.~Felsberg, et~al.
\newblock Eco: Efficient convolution operators for tracking.
\newblock In {\em CVPR}, volume~1, page~3, 2017.

\bibitem{SRDCF}
M.~Danelljan, G.~Hager, F.~Shahbaz~Khan, and M.~Felsberg.
\newblock Learning spatially regularized correlation filters for visual
  tracking.
\newblock In {\em Proceedings of the IEEE International Conference on Computer
  Vision}, pages 4310--4318, 2015.

\bibitem{CCOT}
M.~Danelljan, A.~Robinson, F.~S. Khan, and M.~Felsberg.
\newblock Beyond correlation filters: Learning continuous convolution operators
  for visual tracking.
\newblock In {\em European Conference on Computer Vision}, pages 472--488.
  Springer, 2016.

\bibitem{triSiam}
X.~Dong and J.~Shen.
\newblock Triplet loss in siamese network for object tracking.
\newblock In {\em Proceedings of the European Conference on Computer Vision
  (ECCV)}, pages 459--474, 2018.

\bibitem{PTAV}
H.~Fan and H.~Ling.
\newblock Parallel tracking and verifying: A framework for real-time and high
  accuracy visual tracking.
\newblock In {\em Proc. IEEE Int. Conf. Computer Vision, Venice, Italy}, 2017.

\bibitem{VOT15}
M.~Felsberg, A.~Berg, G.~Hager, J.~Ahlberg, M.~Kristan, J.~Matas, A.~Leonardis,
  L.~Cehovin, G.~Fernandez, T.~Vojir, et~al.
\newblock The thermal infrared visual object tracking vot-tir2015 challenge
  results.
\newblock In {\em Proceedings of the IEEE International Conference on Computer
  Vision Workshops}, pages 76--88, 2015.

\bibitem{BACF}
H.~K. Galoogahi, A.~Fagg, and S.~Lucey.
\newblock Learning background-aware correlation filters for visual tracking.
\newblock In {\em ICCV}, pages 1144--1152, 2017.

\bibitem{CFCF}
E.~Gundogdu and A.~A. Alatan.
\newblock Good features to correlate for visual tracking.
\newblock {\em IEEE Transactions on Image Processing}, 27(5):2526--2540, 2018.

\bibitem{DSiam}
Q.~Guo, W.~Feng, C.~Zhou, R.~Huang, L.~Wan, and S.~Wang.
\newblock Learning dynamic siamese network for visual object tracking.
\newblock In {\em The IEEE International Conference on Computer Vision
  (ICCV).(Oct 2017)}, 2017.

\bibitem{saSiam}
A.~He, C.~Luo, X.~Tian, and W.~Zeng.
\newblock A twofold siamese network for real-time object tracking.
\newblock In {\em Proceedings of the IEEE Conference on Computer Vision and
  Pattern Recognition}, pages 4834--4843, 2018.

\bibitem{ResNet}
K.~He, X.~Zhang, S.~Ren, and J.~Sun.
\newblock Deep residual learning for image recognition.
\newblock In {\em Proceedings of the IEEE conference on computer vision and
  pattern recognition}, pages 770--778, 2016.

\bibitem{GOTURN}
D.~Held, S.~Thrun, and S.~Savarese.
\newblock Learning to track at 100 fps with deep regression networks.
\newblock In {\em European Conference on Computer Vision}, pages 749--765.
  Springer, 2016.

\bibitem{VOT16}
M.~Kristan, A.~Leonardis, and J.~M. et.
\newblock The visual object tracking vot2016 challenge results.
\newblock In {\em Proceedings, European Conference on Computer Vision (ECCV)
  workshops}, pages 777--823, 8Oct. 2016.

\bibitem{VOT17}
M.~Kristan, A.~Leonardis, J.~Matas, and M.~F. et.
\newblock The visual object tracking vot2017 challenge results.
\newblock In {\em 2017 IEEE International Conference on Computer Vision
  Workshop (ICCVW)}, volume~00, pages 1949--1972, Oct. 2017.

\bibitem{AlexNet}
A.~Krizhevsky, I.~Sutskever, and G.~E. Hinton.
\newblock Imagenet classification with deep convolutional neural networks.
\newblock In {\em Advances in neural information processing systems}, pages
  1097--1105, 2012.

\bibitem{SGD}
Y.~LeCun, B.~Boser, J.~S. Denker, D.~Henderson, R.~E. Howard, W.~Hubbard, and
  L.~D. Jackel.
\newblock Backpropagation applied to handwritten zip code recognition.
\newblock {\em Neural computation}, 1(4):541--551, 1989.

\bibitem{siamRPN}
B.~Li, J.~Yan, W.~Wu, Z.~Zhu, and X.~Hu.
\newblock High performance visual tracking with siamese region proposal
  network.
\newblock In {\em Proceedings of the IEEE Conference on Computer Vision and
  Pattern Recognition}, pages 8971--8980, 2018.

\bibitem{LiXi_survey}
X.~Li, W.~Hu, C.~Shen, Z.~Zhang, A.~Dick, and A.~V.~D. Hengel.
\newblock A survey of appearance models in visual object tracking.
\newblock {\em ACM transactions on Intelligent Systems and Technology (TIST)},
  4(4):58, 2013.

\bibitem{SAMF}
Y.~Li and J.~Zhu.
\newblock A scale adaptive kernel correlation filter tracker with feature
  integration.
\newblock In {\em European conference on computer vision}, pages 254--265.
  Springer, 2014.

\bibitem{CSRDCF}
A.~Lukezic, T.~Vojir, L.~C. Zajc, J.~Matas, and M.~Kristan.
\newblock Discriminative correlation filter with channel and spatial
  reliability.
\newblock In {\em CVPR}, volume~6, page~8, 2017.

\bibitem{LDP}
A.~Luke{\v{z}}i{\v{c}}, L.~{\v{C}}. Zajc, and M.~Kristan.
\newblock Deformable parts correlation filters for robust visual tracking.
\newblock {\em IEEE transactions on cybernetics}, 48(6):1849--1861, 2018.

\bibitem{TCNN}
H.~Nam, M.~Baek, and B.~Han.
\newblock Modeling and propagating cnns in a tree structure for visual
  tracking.
\newblock {\em arXiv preprint arXiv:1608.07242}, 2016.

\bibitem{MDNet}
H.~Nam and B.~Han.
\newblock Learning multi-domain convolutional neural networks for visual
  tracking.
\newblock In {\em Proceedings of the IEEE Conference on Computer Vision and
  Pattern Recognition}, pages 4293--4302, 2016.

\bibitem{YTB}
E.~Real, J.~Shlens, S.~Mazzocchi, X.~Pan, and V.~Vanhoucke.
\newblock Youtube-boundingboxes: A large high-precision human-annotated data
  set for object detection in video.
\newblock In {\em Computer Vision and Pattern Recognition (CVPR), 2017 IEEE
  Conference on}, pages 7464--7473. IEEE, 2017.

\bibitem{ImageNet}
O.~Russakovsky, J.~Deng, H.~Su, J.~Krause, S.~Satheesh, S.~Ma, Z.~Huang,
  A.~Karpathy, A.~Khosla, M.~Bernstein, et~al.
\newblock Imagenet large scale visual recognition challenge.
\newblock {\em International Journal of Computer Vision}, 115(3):211--252,
  2015.

\bibitem{VGG}
K.~Simonyan and A.~Zisserman.
\newblock Very deep convolutional networks for large-scale image recognition.
\newblock {\em arXiv preprint arXiv:1409.1556}, 2014.

\bibitem{PAMI_survey}
A.~W. Smeulders, D.~M. Chu, R.~Cucchiara, S.~Calderara, A.~Dehghan, and
  M.~Shah.
\newblock Visual tracking: An experimental survey.
\newblock {\em IEEE Transactions on Pattern Analysis \& Machine Intelligence},
  (1):1, 2013.

\bibitem{VITAL}
Y.~Song, C.~Ma, X.~Wu, L.~Gong, L.~Bao, W.~Zuo, C.~Shen, R.~Lau, and M.-H.
  Yang.
\newblock Vital: Visual tracking via adversarial learning.
\newblock {\em arXiv preprint arXiv:1804.04273}, 2018.

\bibitem{LSART}
C.~Sun, H.~Lu, and M.-H. Yang.
\newblock Learning spatial-aware regressions for visual tracking.
\newblock In {\em IEEE Conf. on Computer Vision and Pattern Recognition
  (CVPR)}, pages 8962--8970, 2018.

\bibitem{Inception}
C.~Szegedy, W.~Liu, Y.~Jia, P.~Sermanet, S.~Reed, D.~Anguelov, D.~Erhan,
  V.~Vanhoucke, and A.~Rabinovich.
\newblock Going deeper with convolutions.
\newblock In {\em Proceedings of the IEEE conference on computer vision and
  pattern recognition}, pages 1--9, 2015.

\bibitem{SINT}
R.~Tao, E.~Gavves, and A.~W. Smeulders.
\newblock Siamese instance search for tracking.
\newblock In {\em Proceedings of the IEEE conference on computer vision and
  pattern recognition}, pages 1420--1429, 2016.

\bibitem{CFNet}
J.~Valmadre, L.~Bertinetto, J.~Henriques, A.~Vedaldi, and P.~H. Torr.
\newblock End-to-end representation learning for correlation filter based
  tracking.
\newblock In {\em Computer Vision and Pattern Recognition (CVPR), 2017 IEEE
  Conference on}, pages 5000--5008. IEEE, 2017.

\bibitem{MLDF}
L.~Wang, W.~Ouyang, X.~Wang, and H.~Lu.
\newblock Visual tracking with fully convolutional networks.
\newblock In {\em Proceedings of the IEEE international conference on computer
  vision}, pages 3119--3127, 2015.

\bibitem{OTB-13}
Y.~Wu, J.~Lim, and M.-H. Yang.
\newblock Online object tracking: A benchmark.
\newblock In {\em Proceedings of the IEEE conference on computer vision and
  pattern recognition}, pages 2411--2418, 2013.

\bibitem{OTB-15}
Y.~Wu, J.~Lim, and M.-H. Yang.
\newblock Object tracking benchmark.
\newblock {\em IEEE Transactions on Pattern Analysis and Machine Intelligence},
  37(9):1834--1848, 2015.

\bibitem{ResNeXt}
S.~Xie, R.~Girshick, P.~Doll{\'a}r, Z.~Tu, and K.~He.
\newblock Aggregated residual transformations for deep neural networks.
\newblock In {\em Computer Vision and Pattern Recognition (CVPR), 2017 IEEE
  Conference on}, pages 5987--5995. IEEE, 2017.

\bibitem{structSiam}
Y.~Zhang, L.~Wang, J.~Qi, D.~Wang, M.~Feng, and H.~Lu.
\newblock Structured siamese network for real-time visual tracking.
\newblock In {\em Proceedings of the European Conference on Computer Vision
  (ECCV)}, pages 351--366, 2018.

\bibitem{EBT}
G.~Zhu, F.~Porikli, and H.~Li.
\newblock Beyond local search: Tracking objects everywhere with
  instance-specific proposals.
\newblock In {\em Proceedings of the IEEE Conference on Computer Vision and
  Pattern Recognition}, pages 943--951, 2016.

\bibitem{DaSiam}
Z.~Zhu, Q.~Wang, B.~Li, W.~Wu, J.~Yan, and W.~Hu.
\newblock Distractor-aware siamese networks for visual object tracking.
\newblock In {\em European Conference on Computer Vision}, pages 103--119.
  Springer, 2018.

\bibitem{zhanglei}
W.~Zuo, X.~Wu, L.~Lin, L.~Zhang, and M.-H. Yang.
\newblock Learning support correlation filters for visual tracking.
\newblock {\em IEEE Transactions on Pattern Analysis and Machine Intelligence},
  2018.

\end{thebibliography}
}

\end{document}